\documentclass{article}

    \PassOptionsToPackage{numbers, compress}{natbib}

    \usepackage[preprint]{neurips_2020}

\usepackage[utf8]{inputenc} 
\usepackage[T1]{fontenc}    
\usepackage{hyperref}       
\usepackage{url}            
\usepackage{booktabs}       
\usepackage{amsfonts}       
\usepackage{nicefrac}       
\usepackage{microtype}      
\usepackage{algorithm}
\usepackage{pifont}
\usepackage{multirow}
\usepackage{arydshln}
\usepackage{etoolbox}
\usepackage{algorithmic}
\usepackage{graphicx}%
\usepackage{bbm}

\usepackage{multicol}
\usepackage{listings}
\usepackage{times}
\usepackage{epsfig}
\usepackage{graphicx}
\usepackage{subcaption}
\usepackage[font=small]{caption}
\usepackage{mwe}
\usepackage{amsmath}
\usepackage{amssymb}
\usepackage{booktabs}
\usepackage{algorithm}
\usepackage{algorithmic}
\usepackage{appendix}
\usepackage{amsmath,amssymb}

\usepackage{textcomp}

\usepackage{mathtools}
\usepackage{float}
\usepackage{graphicx}%
\usepackage{bbm}
\usepackage{hyperref}
\usepackage{multicol}
\usepackage{listings}
\usepackage[table]{xcolor}
\usepackage{graphicx}
\usepackage{caption}
\usepackage{subcaption}
\usepackage[thinlines]{easytable}
\usepackage{systeme}

\pdfoutput=1

\title{Physically Feasible Vehicle Trajectory Prediction}

\author{%
}

\author{Harshayu Girase$^{\dagger\S*}$ \;\; Jerrick Hoang$^{\dagger*}$ \;\; Sai Yalamanchi$^{\dagger}$ \;\; Micol Marchetti-Bowick$^{\dagger}$ \;\; \vspace{10pt}\\ 
	$^\dagger$ Uber ATG  \;\;  
	$^\S$ UC Berkeley \;\;  
	}

\newcommand\blfootnote[1]{%
  \begingroup
  \renewcommand\thefootnote{}\footnote{#1}%
  \addtocounter{footnote}{-1}%
  \endgroup
}

\begin{document}

\maketitle

\begin{abstract}
    Predicting the future motion of actors in a traffic scene is a crucial part of any autonomous driving system. Recent research in this area has focused on trajectory prediction approaches that optimize standard trajectory error metrics. In this work, we describe three important properties -- physical realism guarantees, system maintainability, and sample efficiency -- which we believe are equally important for developing a self-driving system that can operate safely and practically in the real world. Furthermore, we introduce \textbf{PTNet} (PathTrackingNet), a novel approach for vehicle trajectory prediction that is a hybrid of the classical pure pursuit path tracking algorithm and modern graph-based neural networks. By combining a structured robotics technique with a flexible learning approach, we are able to produce a system that not only achieves the same level of performance as other state-of-the-art methods on traditional trajectory error metrics, but also provides strong guarantees about the physical realism of the predicted trajectories while requiring half the amount of data. We believe focusing on this new class of hybrid approaches is an useful direction for developing and maintaining a safety-critical autonomous driving system.
\end{abstract}
\vspace{-.1cm}
\section{Introduction}
\vspace{-1mm}

Over the past several years, significant advancements have been made in autonomous driving. One active area of research is in methods for predicting the future motion of surrounding actors in the scene. In this work, we focus on the prediction problem, and in particular, 
we want to direct the attention of the research community to a less explored approach: combining robotics with machine learning. We argue that this hybrid approach is useful in order to maintain a safety critical system over a long period of time. In particular, we focus on the following three aspects. 

\begin{figure}
\centering
\begin{subfigure}{.45\textwidth}
  \centering
  \includegraphics[width=1\linewidth, trim=0 1.6cm 0 0, clip]{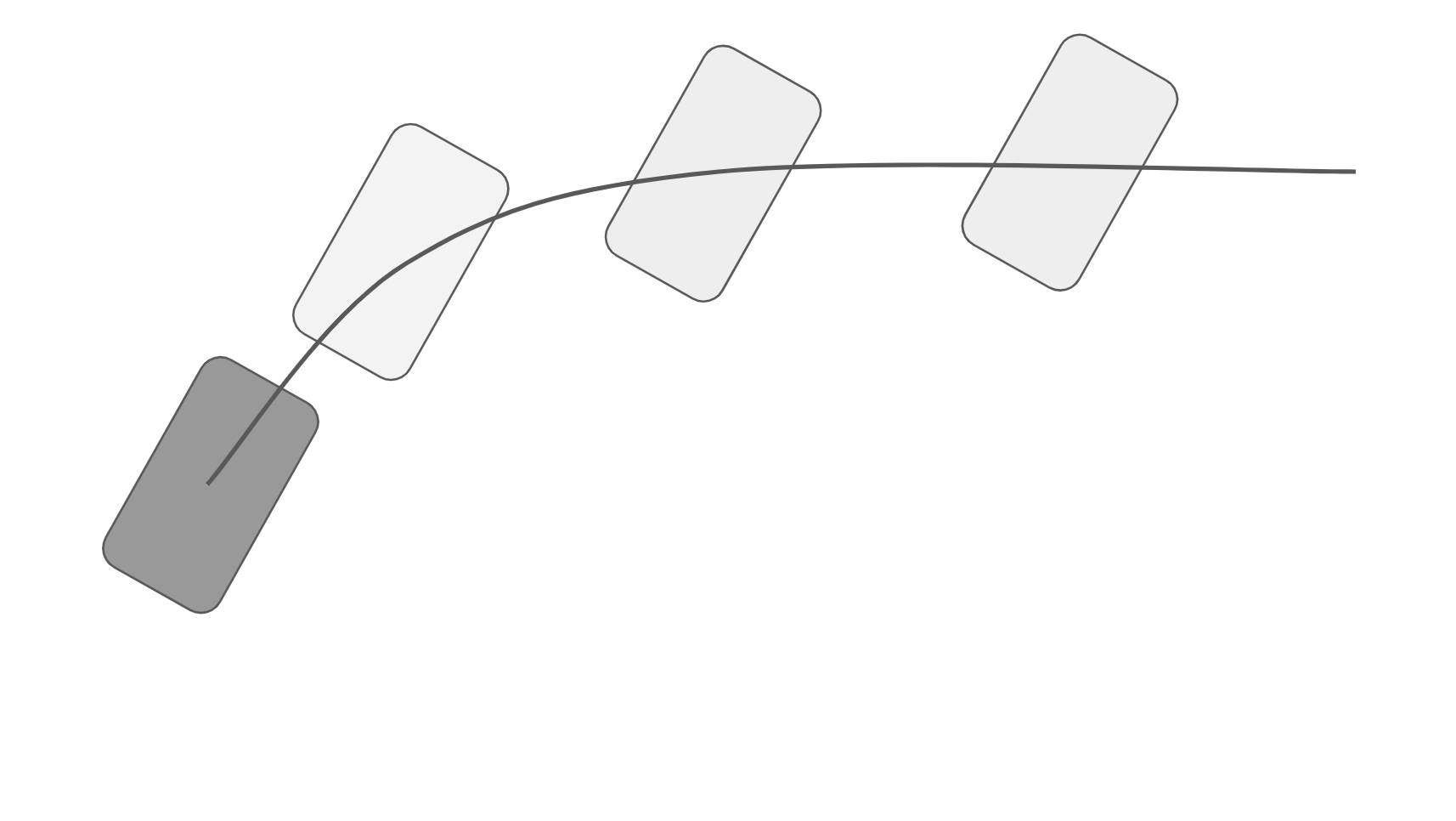}
  \caption{A hypothetical scenario where the predicted trajectory has zero displacement error. However, the predicted orientation makes the trajectory dynamically infeasible for a vehicle to follow.}
  \label{fig:sub1}
\end{subfigure}%
\hfill
\begin{subfigure}{.5\textwidth}
  \centering
  \includegraphics[width=1\linewidth]{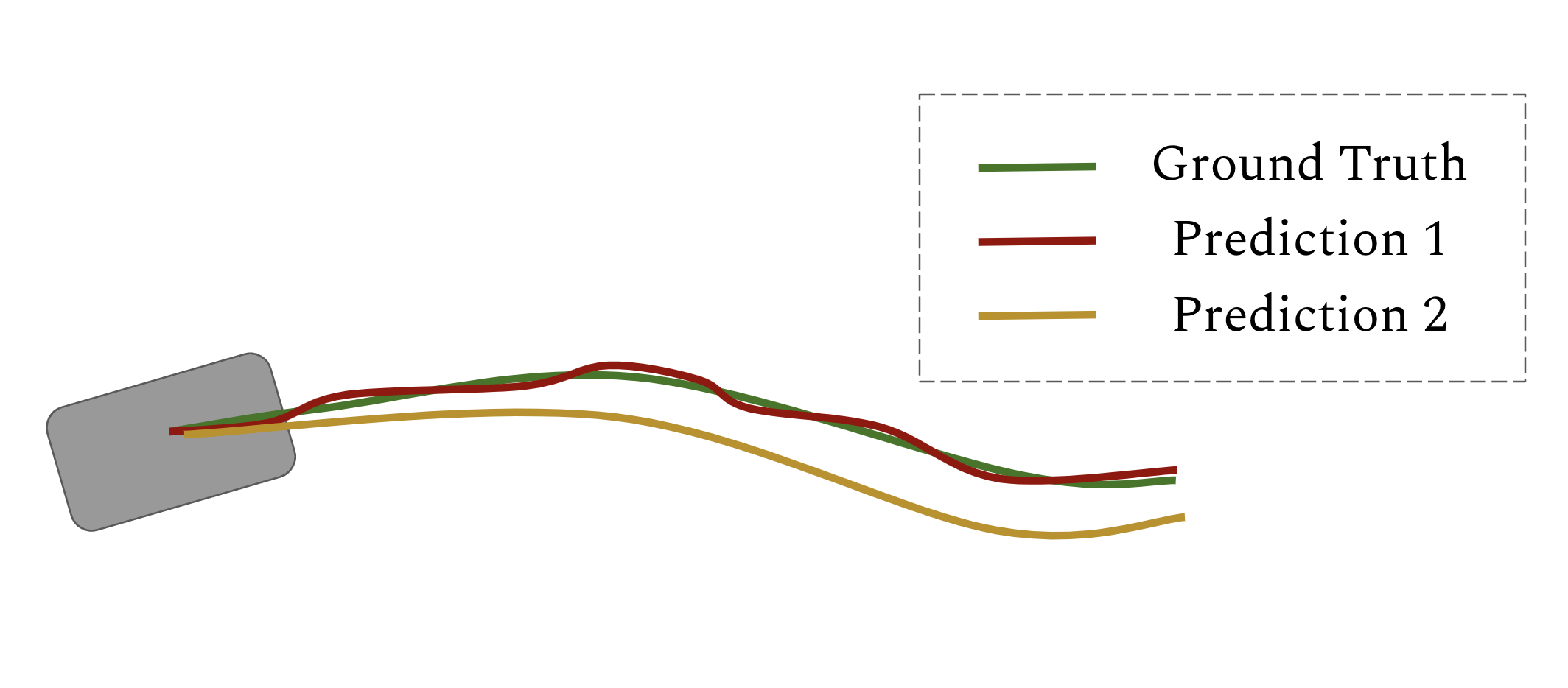}
  \caption{A hypothetical scenario where prediction 1 (red) has smaller trajectory error than prediction 2 (yellow) in comparison to ground truth (green). However, the red predicted trajectory is physically unrealistic.}
  \label{fig:sub2}
\end{subfigure}
\caption{Two scenarios scenarios demonstrating that trajectory error metrics do not fully capture the important characteristics of the predicted trajectories.}
\label{fig:motivation}
\vspace{-0.5cm}
\end{figure}

\textbf{Physical realism guarantees}: Despite being a very powerful toolkit, pure machine learning methods provide few theoretical guarantees with regard to the physical dynamics of the predicted trajectories. On the other hand, robotic techniques for understanding rigid body movement have been extensively studied and theoretically understood \cite{9179748, sontag2013mathematical}. In this paper, we propose a hybrid technique, leveraging the power of learning from data along with the advantage of strong theoretical guarantees that robotic techniques provide. We also emphasize the importance of evaluating prediction quality based on trajectory feasibility metrics in addition to trajectory error metrics. Figure \ref{fig:motivation} shows two hypothetical situations where trajectory error metrics do not paint a full picture. We show quantitatively and qualitatively that our model outperforms state-of-the-art models on key feasibility metrics.

\textbf{System maintainability}: Maintainability is the degree to which an application is understood, or the ease with which an application can be maintained, enhanced, and improved over time. 
Using the framework of \textit{technical debt}, the authors of \cite{43146} argue that despite the quick wins machine learning\blfootnote{$*$ indicates equal contribution} methods bring to a system, it is remarkably easy to incur a massive maintenance cost over the long run. Specifically, the authors emphasize boundary erosion, arguing that machine learning models create entanglement which make isolated improvements impossible. Combining robotic techniques with machine learning naturally introduces stronger abstraction boundaries, creating an encapsulated and modular design that leads to more maintainable code in the long run \cite{fowler2018refactoring}.

\textbf{Sample efficiency}: It is expensive to collect a large amount of real-world driving data, particularly if the autonomous driving system also depends on having a high-definition map. Machine learning methods require a large amount of data in order to learn high-dimensional correlations from scratch. Robotic techniques, on the other hand, already incorporate distilled knowledge about the world in the form of physics equations. A hybrid approach increases sample efficiency while still learning from data. We show that our model is twice as sample efficient and still outperforms or matches state-of-the-art performance in trajectory error by explicitly modeling and executing motion profiles.

Overall, in order to develop and maintain a safety critical autonomous driving system, we believe that it is useful to explore more structured machine learning approaches by combining them with existing and well-studied robotics techniques. In this work, we combine graph neural networks with the Pure Pursuit path tracking algorithm \cite{coulter1992implementation}, leveraging the power of data to learn a motion profile predictor for each actor and executing the motion profile with a Pure Pursuit model. We show quantitatively and qualitatively that our method achieves the same level of performance as state-of-the-art methods, while also providing strong physical realism guarantees and improving on sample efficiency.

\vspace{-.1cm}
\section{Related Work}
\vspace{-1mm}

With the recent success of deep learning in many domains, recent work has focused on deep neural network based approaches for trajectory prediction. Specifically, the recurrent neural network (RNN) has shown promising performance for sequence learning and many works have employed long short-term memory (LSTM) and gated recurrent unit (GRU) networks for future motion forecasting given past sequential observations \cite{sutskever2014sequence, lee2017desire, kim2017probabilistic, ma2019trafficpredict, watters2017visual}. In order to capture surrounding information, recent papers have proposed to use contextual scene information for scene-compliant prediction \cite{cui2019multimodal, djuric2020uncertainty, bansal2018chauffeurnet, zhao2020tnt}. Most of these methods encode scene context features using convolutional neural networks (CNN) and then use learnt embeddings downstream to predict actor trajectories. Another family of deep-learning-based approaches has explored graph neural networks to model interactions between agents \cite{huang2019stgat, casas2019spatially, zhang2019stochastic, lee2019joint, mohamed2020social, kumar2020interaction}. These deep networks learn vehicle dynamics, such as feasible steering and acceleration, from the large number of example trajectories available during training. However, unconstrained motion prediction may violate the dynamic feasibility constraints of vehicles. Most of these methods only predict the center positions of a vehicle or independently predict \cite{casas2018intentnet, cui2020deep} position and heading which may still result in infeasible discrepancies between the two, and as a result may poorly capture the vehicle's occupancy in space.

Other approaches also introduce structure into the task by exploring goal-conditioned ideas which model actor intent and goals prior to predicting the trajectory. Rhinehart et al.~\cite{rhinehart2019precog} propose PRECOG, a goal-conditioned approach for multi-agent prediction. More recently, Mangalam et al.~\cite{mangalam2020not, mangalam2020goals} propose an endpoint conditioned prediction scheme which conditions pedestrian predictions on goal destinations. In the vehicle setting, GoalNet \cite{zhang2020map} uses lane centerlines as goal paths and predicts trajectories in the path-relative Frenet frame. The authors of MultiPath \cite{chai2020multipath} similarly introduce the idea of trajectory anchors to provide more structure for the problem. While in practice these structured approaches could provide more system maintainability and sample efficiency, they still do not provide an absolute guarantee on the dynamic feasibility of the predicted trajectories since the predicted offsets could be arbitrary. 
Our method leverages the goal-based trajectory prediction approach introduced in \cite{zhang2020map}, which allows long-term behavior to be captured. However, we also incorporate a path tracking algorithm that provides strong guarantee on the physical realism of the resulting trajectories. 

There have been a few prior approaches that combine robotics techniques with machine learning. These methods have focused primarily on state estimation or short-term instantaneous trajectory prediction \cite{kalman1960new, chen2011kalman} through kinematic-based vehicle models such as a Kalman Filter. While these models work well for short-term prediction, they become increasingly unreliable for longer-term prediction as the vehicle will change control inputs based on scene elements or other vehicles. The majority of robotics-focused research that has been applied to the autonomous driving domain has been on the motion planning side, e.g.~\cite{howard2010receding, kong2015kinematic, bansal2020combining}. In contrast, our approach combines learning and robotics for the prediction task. CoverNet \cite{phan2020covernet} is a recent approach that ensures physical realism by turning the trajectory prediction problem into a classification task over a set of dynamically feasible trajectories. However, in order to achieve good coverage over the space of possible future motion, a large number of trajectories must be generated, which might be impractical for a system that must run in real time. 
The deep kinematic model (DKM) introduced by Cui et al.~\cite{cui2020deep} is closest to our philosophy, as they predict control inputs such as acceleration and steering prior to trajectory roll-out. While they ensure dynamic feasibility of their trajectories, their proposed method does not explicitly use goal path information to guide predictions that may be useful for longer-horizon predictions. Our work relies on Pure Pursuit path tracking \cite{coulter1992implementation} to produce goal-directed trajectories that yield better long-term predictions while still providing strong guarantees on curvature and higher-order dynamics.
\vspace{-.1cm}
\section{Method}
\vspace{-1mm}

In this section, we present our method for combining a flexible data-driven model with a structured path tracking algorithm to produce dynamically feasible trajectory predictions for vehicles. Our approach is built on top of GoalNet, a goal-based trajectory prediction model introduced in \cite{zhang2020map}. The key difference between our model and GoalNet is that we replace the final trajectory prediction layer with a highly structured Pure Pursuit path tracking (PT) layer. We therefore call our model PTNet.

In this work, we focus solely on the trajectory prediction problem. We assume that there is a real-time detection and tracking system onboard the SDV to detect and estimate the states of surrounding traffic actors. For each actor, our model is given the actor's dynamics state (current heading, position, velocity, and acceleration) and its past positions. Furthermore, mapped lane boundaries and traffic sign locations are also available as inputs. Given this information, our model then makes multi-modal spatio-temporal predictions of the actor's positions over the next $T$ timesteps.

\begin{figure}
    \centering
    \includegraphics[width=\linewidth,trim={0cm 2.2cm 0cm .4cm},clip]{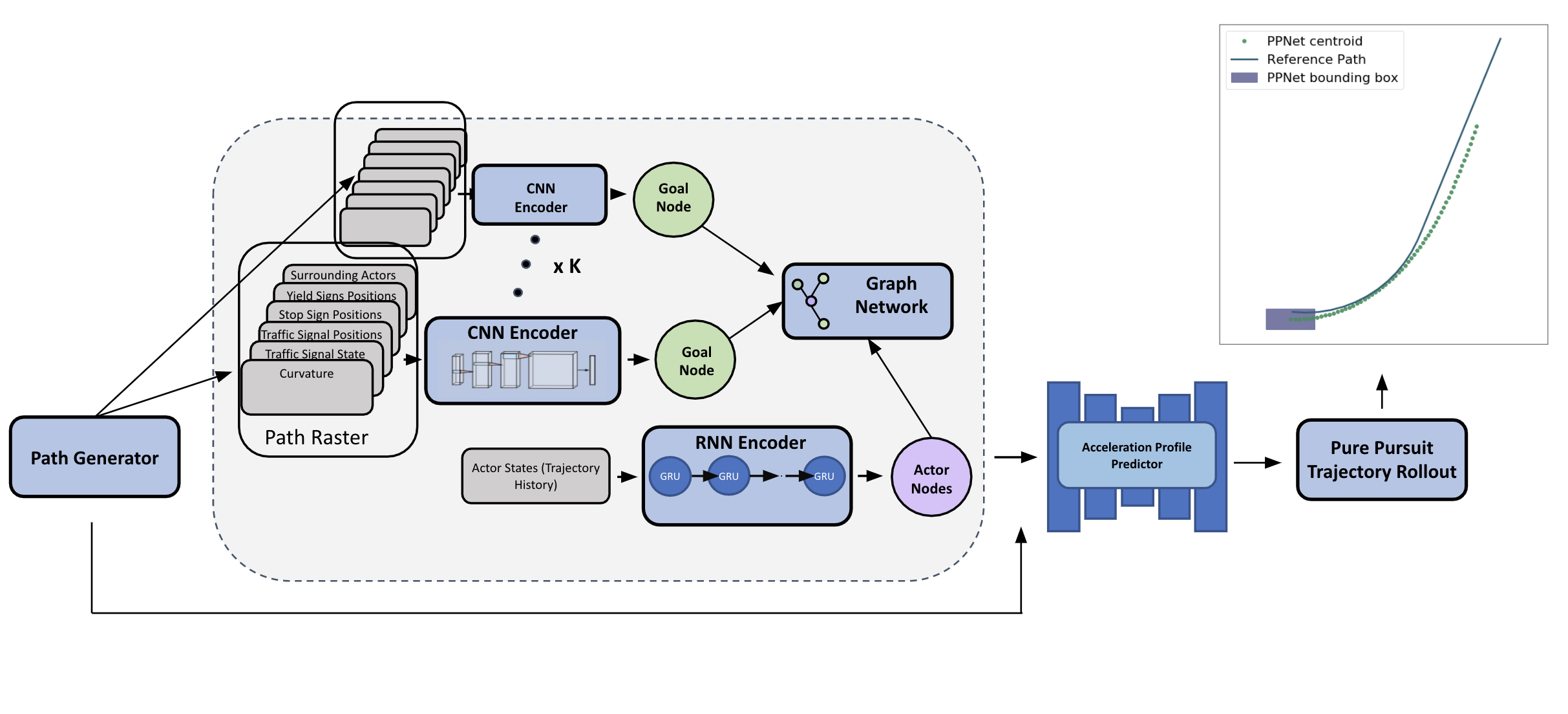}
    \vspace{-.2cm}
    \caption{An overview of our method. Given a high-definition map and a perception system that outputs tracked objects, our method consists of three main steps. (1) Path generation: for each object, we generate a set of goal paths. (2) Input encoding: for each actor, we construct a graph that consists of one single actor node and a variable set of goal nodes, and then encode the actor state and the goal information into latent features using a GNN. (3) Trajectory development: for each goal, we predict $N$ temporal motion profiles and execute each motion profile using the PurePursuit path tracking algorithm to produce a trajectory. }
    \label{fig:method}
\end{figure}

\subsection{Background}

Our method is built on the path generation and graph encoder layers introduced in GoalNet \cite{zhang2020map} combined with the Pure Pursuit algorithm \cite{coulter1992implementation}. In the remainder of this section, we provide some background on each of these building blocks.

\subsubsection{Path Generation}
\vspace{-1mm}

The path generation module generates a set of map-based goal paths and one additional map-free path for each actor. To generate map-based paths, we treat the map as a lane graph, where each node is a lane (with no branching points) and a directed edge connecting two nodes if the source node is the predecessor lane of the target node. Given the lane graph and the actor's current position, we query for the closest nodes (we use a radius $r = 2$m) and roll out the paths by traversing the graph and connecting the centerlines of the lanes. The result is a list of lane sequences we call \textit{map-based paths}. Since the map might not capture the high-level intention of the actor, we also generate one additional \textit{map-free path} in the direction of travel of the actor. For each actor $A^i$, we then transform each goal path $G^{i}_j$ into the actor's frame of reference. These will later be used as reference paths for our path tracking algorithm.

%

\subsubsection{Graph Networks Encoder}
\vspace{-1mm}

To handle a variable number of goals for each actor, we use a graph network \cite{scarselli2008graph} following the method introduced in GoalNet \cite{zhang2020map}. In particular, we construct a mini graph for each actor. The graph consists of a variable number of goal nodes and a single actor node. The actor's current and past states are encoded into the initial actor node features. The path and context information along each path, e.g.~signage locations and curvature, is encoded into initial goal node features. For the initial set of features for each edge corresponding to a goal, we use the actor's current velocity and acceleration to construct a 0-jerk rollout and project it to onto the path.  We then follow the equation proposed in \cite{zhang2020map} for our graph network updates.  

Specifically, for each actor $i$, denote $v^{\ell}_i$ to be the encoded actor node attributes at layer $\ell$. Let the set of all goals for each $i$ be $G_i$. For each goal $j \in G_i$ of actor $i$, denote $e^{\ell}_{ij}$ to be the edge attributes at layer $\ell$ and let $g_j$ be the goal node attributes for the goal $j$. Note that $g_j$ does not have a layer index since the goal nodes do not have incoming edges so they are not updated at each iteration. The update equations are given by,
\begin{align}
    e^{\ell+1}_{ij} &= \phi_e(v^{\ell}_i, e^{\ell}_{ij}, g_j) \\
    v^{\ell+1}_i &= \phi_v(v^{\ell}_i, \psi(\{ e^{\ell}_{ij} \}_{j \in G_i}))
\end{align}
We use a 2-layer MLP for $\phi_v$ and $\phi_a$ and the mean function for $\psi$. This simple graph network contains a total of 2 layers.

\subsubsection{Pure Pursuit Algorithm}
\vspace{-1mm}


\begin{figure}
    \centering
    \includegraphics[width=0.9\linewidth,trim={0cm .3cm 0cm .4cm},clip]{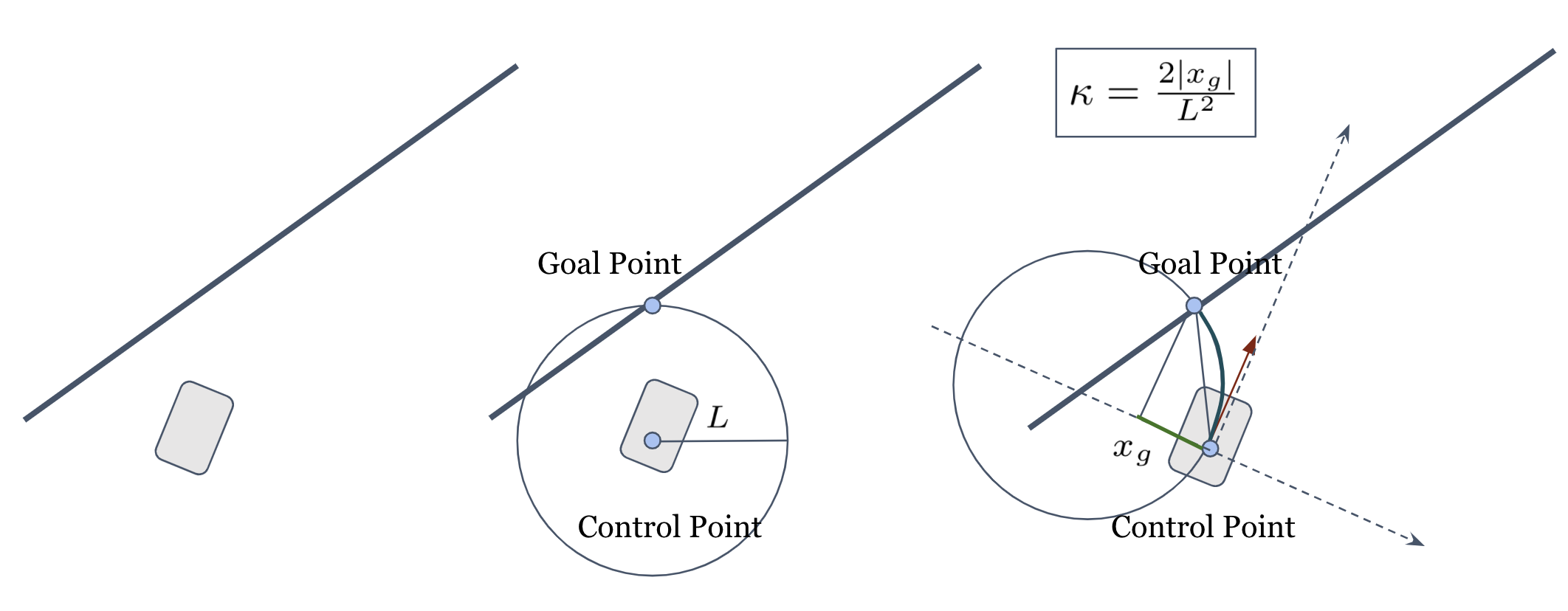}
    \caption{An overview of the Pure Pursuit path tracking update. }
    \label{fig:purepursuit}
\end{figure}

Pure Pursuit \cite{coulter1992implementation} is a simple path tracking algorithm that calculates the necessary arc to reach a "goal point" on the path. The algorithm is aptly named based on the way humans drive -- we look at a point we want to go to and control the car to reach that point. The high level steps of the Pure Pursuit algorithm are outlined in Figure \ref{fig:purepursuit} and are described in detail below. 

\textbf{Finding the target point on the path}: Given the actor's position and a goal path, the algorithm first tries to find a target point on the path to track. In particular, given a fixed lookahead distance $L$, we find all possible intersections of the circle of radius $L$ centering at the actor's control point with the goal path. Our paths are generated so that they are sufficiently close to the actor such that an initial intersection point always exists. In the case where there is more than one intersection, we only consider the point ahead of the actor on the path. In all of our experiments, we use $L = 10$m.

\textbf{Assume circular motion and calculate curvature}: Assuming circular motion, we find the circle passing through the goal point and tangent to the control point in the direction of heading. Given the geometric setup, it can be proved (see \cite{coulter1992implementation}) that the curvature can be obtained by $\kappa = \nicefrac{2|x_g|}{L^2}$ where $x_g$ is the x-component of the goal point in the actor's frame. However, this curvature can be infeasible. To ensure that we only execute feasible motion, we calculate the final curvature by $\kappa = \min(2|x_g|/L^2, M_c)$ where $M_c$ is the maximum possible curvature that can be executed by a vehicle. In our experiments, we choose $M_c = 0.3$. 

\textbf{Update path tracking state}: Assume the state is comprised of position $p^i_{t} = (x^i_t, y^i_t) \in \mathbb{R}^2$, velocity $v^i_{t} \in \mathbb{R}^2$, and heading $h^i_{t} \in [-\pi, \pi)$. We use the following state update equations. The input acceleration at each time step is predicted by the acceleration profile predictor described in the previous section.
\[
\systeme*{
 x^{i}_{t+1} = x^i_{t} + \cos(h^i_{t})v^i_{t}\Delta_t,
 y^{i}_{t+1} = y^i_{t} + \sin(h^i_{t})v^i_{t}\Delta_t,
 v^{i}_{t+1} = v^{i}_{t} + a^{i}_{t}\Delta_t, 
\kappa = \min(2x_g/L^2; M_{\kappa}), 
 h^i_{t+1} = h^i_{t}  + v^i_{t}\Delta_t \kappa,}
\]

\subsection{PTNet}

Our main contribution is connecting the idea of map-adaptive reference path generation and scoring introduced by \cite{zhang2020map} with a structured path following algorithm introduced by \cite{coulter1992implementation}. The outline of our method is described in Figure \ref{fig:method}. The first two components, path generation and graph network encoding, are taken from GoalNet. The last component is a differentiable Pure Pursuit layer. We connect these two components using a temporal profile predictor. This layer also serves as a semantically meaningful abstraction between the two components. Specifically, the output of this layer is a sequence of acceleration values over time which can be visualized and has a specific physical meaning, unlike most intermediate layers in a fully learned system. We implemented all operations such that the system remains end-to-end trainable. The predicted acceleration profile as well as curvature update in PurePursuit provides theoretical guarantees on physical realism. Also, by directly leveraging a motion model, we provide a starting point from which the model can learn, making it more sample efficient.

\subsubsection{Multi-Modal Acceleration Profile Generation}
\vspace{-1mm}

For each of our spatial modes (anchored by reference paths), we generate $N$ different temporal modes. Temporal modes can simply be modeled by generating multiple different acceleration profiles to be inputted into our path tracking algorithm. We follow the same unsupervised training scheme proposed in \cite{cui2019multimodal} to learn different acceleration profile modes. In particular, we have $N$ acceleration prediction networks, each learning a different motion profile, and a mode prediction network to learn the probabilities for predicting each mode. On each iteration, only the mode whose output is closest to the ground truth is penalized in the loss. To ensure physical realism, we constrain the output of each network to be within -8 $m$/$s^2$ and 8 $m$/$s^2$ by using a scaled $\tanh$ activation. The acceleration prediction layer serves as a strong abstraction between input encoding and path tracking layer, which enforces system maintainability. By predicting in control space as an intermediate step instead of predicting position directly, the model also provides a semantically meaningful intermediate layer which improves interpretability.

\subsubsection{Loss Function}
\vspace{-1mm}

The loss function used to train this model is a combination of the mode classification loss and trajectory regression loss. For each actor $i$, we identify the set of goals $\mathcal{G^*}_i$ that match the future ground truth trajectory of the actor (we use the same path labeling algorithm defined in \cite{zhang2020map}). For each actor $i$, we define the target probability of $1/|\mathcal{G^*}_i|$ for each goal. For each matching goal, we assign target probability of 1 for the best matching temporal mode. The target probability of a trajectory mode is the product of the target probability of the spatial (goal) mode and the target probability of the temporal mode (one-hot). For each actor we also calculate the $L1$ smooth loss between the ground-truth trajectory and the trajectory modes weighted by probability,

\begin{align*}
    \mathcal{L} = \sum_{i \in \mathcal{A}} [\underbrace{(-\sum_{m \in \mathcal{T}^i} p_m \log{\hat{p}_m})}_{\text{Mode Classification Loss}} + \underbrace{(\sum_{m \in \mathcal{T}^i} p_m ||\tau_i - \hat{\tau}_m||_1)}_{\text{Trajectory Error Loss}}]
\end{align*}

Here we denote the set of all actors to be $\mathcal{A}$ and the set of all spatio-temporal trajectory modes for an actor $i$ to be $\mathcal{T}^i$. We define the target probability mass for mode $m$ to be $p_m$ and the predicted mass to be $\hat{p}_m$. We denote the predicted trajectory for temporal mode $m$ to be $\hat{\tau}_m$ and the ground truth trajectory of actor $i$ to be $\tau_i$.
\vspace{-.1cm}
\section{Experiments}
\vspace{-1mm}

\subsection{Datasets}
\vspace{-1mm}

We evaluate our method on two datasets: our internal dataset and the public NuScenes dataset \cite{nuscenes2019}. The NuScenes dataset has 1.4M objects over 40K frames collected from 15 hours of driving in Boston and Singapore. Our internal dataset has 138M objects (60\% of which are vehicles) over 6M frames and was collected from various cities in the United States. We only consider non-parked vehicles and those for which we observe at least 6 seconds of future.

\subsection{Baselines}
\vspace{-1mm}

We compare our model with 4 different baselines: GoalNet \cite{zhang2020map}, Multiple Trajectory Prediction (MTP) \cite{cui2019multimodal}, MultiPath \cite{chai2020multipath} and CoverNet \cite{phan2020covernet}. For CoverNet, we use the static version of CoverNet which relies on having a predefined set of trajectories. We directly use the publicly released set of 2206 trajectories for the NuScenes dataset. We did not compare to CoverNet on our internal dataset. MTP predicts a fixed number of different modes and encourages diversity by only updating the winning mode. In our experiments, we use 3 modes for MTP. MultiPath uses a fixed-size set of spatial-temporary trajectory anchors which are estimated by running a clustering algorithm on the training dataset. The method then makes trajectory predictions by outputting offsets from the best anchor. We use 64 modes for our MultiPath comparison. 

\subsection{Feasibility Metrics}
\label{sec:feasibility-metrics}
\vspace{-1mm}

In addition to evaluating our method and baselines on standard trajectory error metrics, we also evaluate them on other metrics that capture physical realism. To define the metrics, we first separate the concept of heading into \textit{bounding box heading} and \textit{motion heading}. An actor's bounding box heading is the direction the actor is facing. In contrast, motion heading is defined to be the direction between two consecutive control points. For our model, these concepts are the same. However, we want to define a set of metrics that can be broadly applicable. Next, we define four key directions that will be used to compute the feasibility metrics. The longitudinal direction is defined to be the direction of the bounding box heading. The lateral direction is defined to be the direction orthogonal to the longitudinal direction. The {traversal} direction is defined to be the direction of the motion heading. The {centripetal} direction is defined to be the direction orthogonal to the traversal direction. Using these concepts, we define a set of physical feasibility metrics below.

\textbf{Curvature Violation}: The curvature over a trajectory segment is given by $\kappa = \nicefrac{2\sin \Delta h^i}{\Delta p^i}$, where $\Delta h^i$ denotes the change in bounding box heading and $\Delta p^i$ denotes the change in position between two waypoints. A trajectory contains a curvature violation if $\kappa >$ 0.3 m$^{-1}$ at any point along the trajectory.

\textbf{Lateral Speed Violation}: If we decompose the speed vector $v^i_t$ into its lateral and longitudinal components, a trajectory contains a lateral speed violation if at any given point $t$, the instantaneous lateral speed is greater than 1 m$/$s.

\textbf{Centripetal Acceleration Violation}: If we decompose the acceleration vector $a^i_t$ into its traversal and centripetal components, a trajectory contains a centripetal acceleration violation if at any given point $t$, the instantaneous centripetal acceleration is greater than 10 m$/$s$^2$.

\textbf{Traversal Acceleration Violation}: If we decompose the acceleration vector $a^i_t$ in to its traversal and centripetal components, a trajectory contains a traversal acceleration violation if at any given point $t$, the instantaneous centripetal acceleration is smaller than $-12$ m$/$s$^2$ or greater than 8 m$/$s$^2$.

All of the above boundary constraints are chosen from studying the physical constraints (acceleration, turning radius) of a standard mid-size SUV. We also made sure that none of the ground truth trajectories in our datasets violate any of these constraints. 

\subsection{Results}
\vspace{-1mm}

We first compare our method with all baselines on a set of general trajectory error metrics. For PTNet and GoalNet, we use the suffix "-$N$T" to refer to a model with $N$ temporal modes. 
In particular, we report the cross-track (CTE), along-track (ATE) and displacement error (DE) on the most probable trajectory in both our internal dataset and NuScenes dataset. The results are shown in Table \ref{tab:results-trajectory-error}. In this table, the errors are averaged over all horizons. The precise definitions of these metrics can be found in the Appendix. We observe that PTNet performs on par with GoalNet on these trajectory error metrics while beating all other baselines across all metrics. 

\begin{table}
\centering
\subfloat[Comparison of the most probable trajectory error on the NuScenes dataset.]{%
\begin{tabular}{lccc}
\toprule
Method & avg ATE  & avg CTE  & avg DE \\
\midrule
PTNet-1T & \textbf{1.82} & 0.51 & \textbf{2.04}  \\
GoalNet-1T & 1.84 & \textbf{0.48} & 2.05  \\
MTP & 2.44 & 0.77 & 2.77  \\
MultiPath & 3.62 & 0.90 & 4.01  \\
CoverNet & 4.07 & 1.14 & 4.57  \\
\end{tabular}}%
~~~~~~
\subfloat[Comparison of the most probable trajectory error on our internal dataset. CoverNet results are not available since CoverNet requires a trajectory anchor set that is specific to the dataset.]{%
\begin{tabular}{lccc}
\toprule
Method & avg ATE & avg CTE  & avg DE \\
\midrule
PTNet-1T & \textbf{2.15} & 0.49 & 2.38  \\
GoalNet-1T & 2.16 & \textbf{0.47} & \textbf{2.37}  \\
MTP & 2.39 & 0.59 & 2.67  \\
MultiPath & 2.59 & 0.66 & 2.91  \\
CoverNet & --- & --- & ---  \\
\end{tabular}}
\caption{Trajectory error metrics comparison on the public NuScenes dataset and our internal dataset.}
\label{tab:results-trajectory-error}
\end{table}

\begin{figure}
\centering
\begin{minipage}{\textwidth}
\centering
{\includegraphics[width=.45\linewidth,clip]{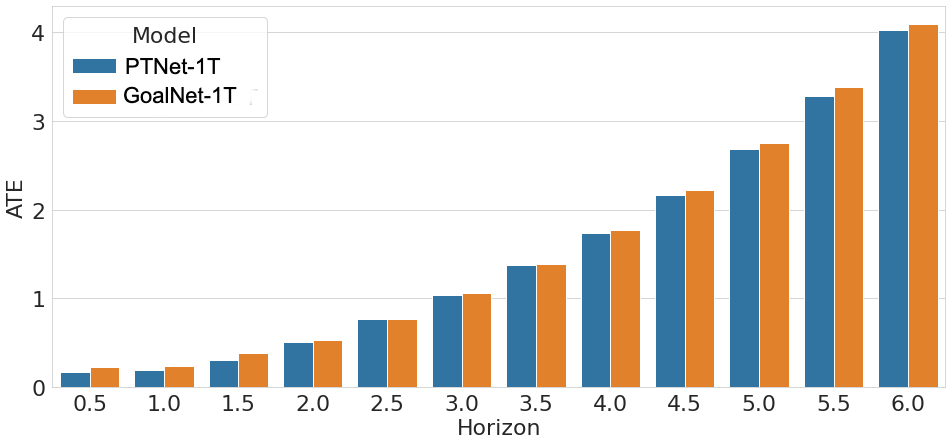}}
{\includegraphics[width=.45\linewidth,clip]{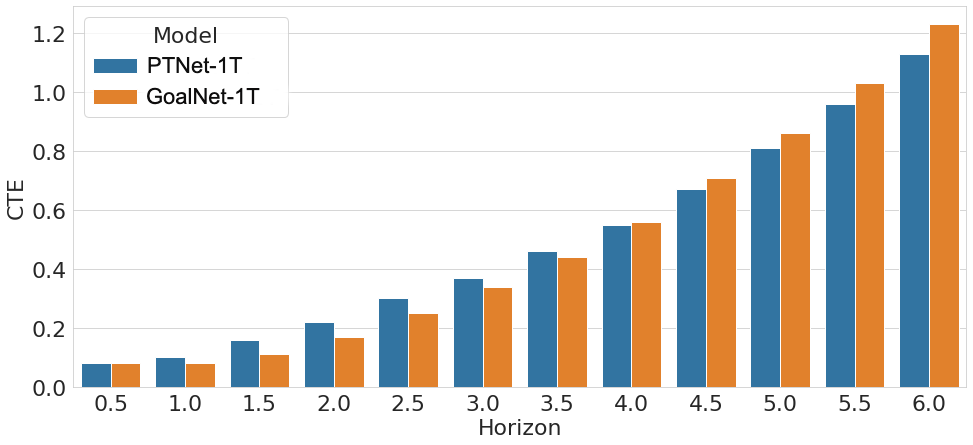}} 
\label{}
\end{minipage}
\label{fig:my_label}
\caption{Trajectory error vs horizon comparison between PTNet-1T and GoalNet-1T. PTNet outperforms GoalNet-1T in along-track error due to its ability to explicitly model the acceleration profile for each goal.}
\label{fig:horizon}
\end{figure}

Next, we dive deeper into comparing PTNet-1T and GoalNet-1T. Specifically, we report the along-track and cross-track errors of the predicted trajectory mode that best matches the ground truth against the prediction horizon. The results are shown in Figure \ref{fig:horizon}. We observe that on best matching error metrics, PTNet-1T outperforms GoalNet-1T on along-track error across all horizons. This could be attributed to PTNet's explicit representation of the acceleration profile, which makes the task easier to learn. On cross-track error, PTNet outperforms GoalNet in later horizons (> 3.5 seconds) but is worse in shorter horizons (< 3.5 seconds). In general, the short-term behavior of an actor is much easier to model with fewer constraints, while long-term behavior is easier to model with more structure. This is why we posit incorporating the structure from a robotics-based approach helps PTNet outperform GoalNet in the longer horizons.

\begin{figure}

\begin{minipage}{\textwidth}
\centering
{\includegraphics[width=.24\linewidth,clip]{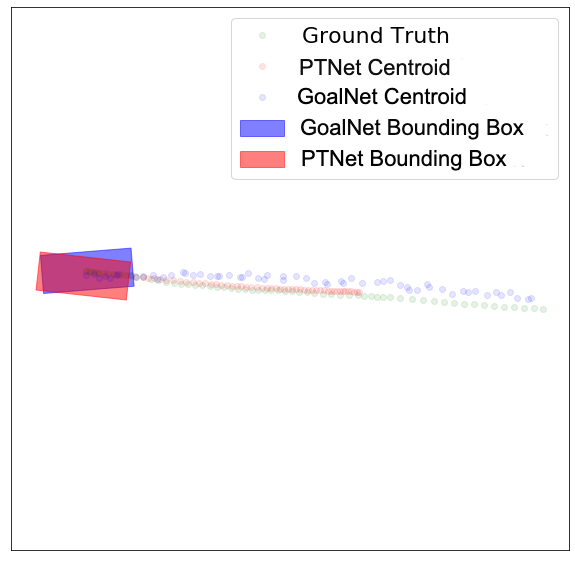}}
{\includegraphics[width=.24\linewidth,clip]{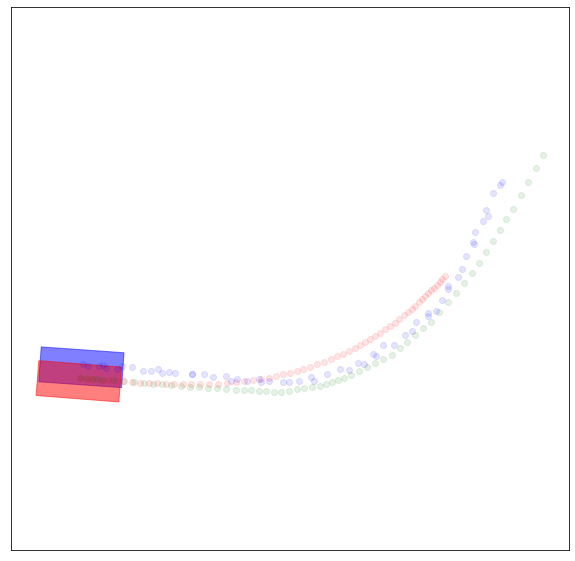}} 
{\includegraphics[width=.24\linewidth,clip]{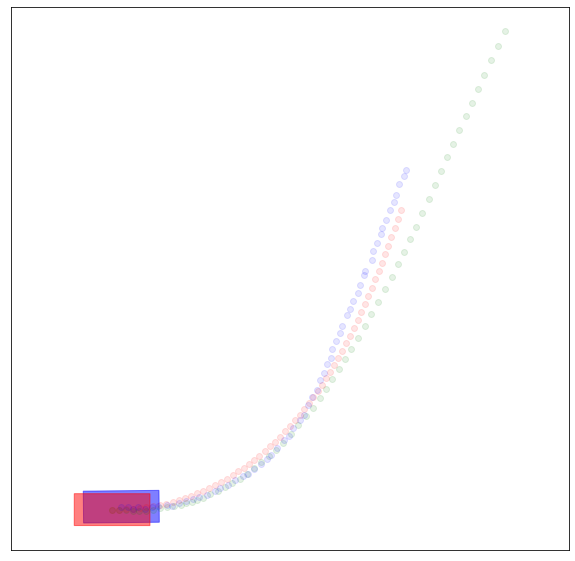}}
{\includegraphics[width=.24\linewidth,clip]{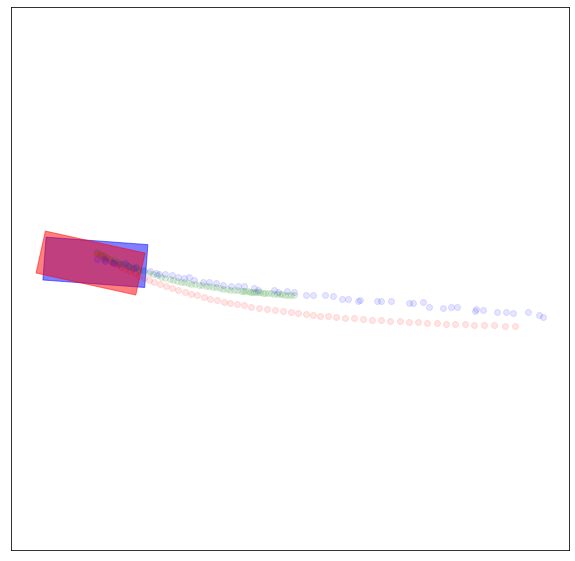}} 
\end{minipage}
\begin{minipage}{\textwidth}
\centering
{\includegraphics[width=.24\linewidth,clip]{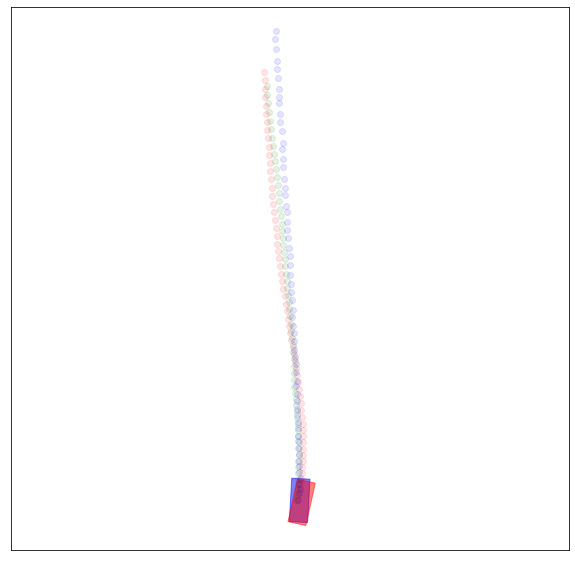}}
{\includegraphics[width=.24\linewidth,clip]{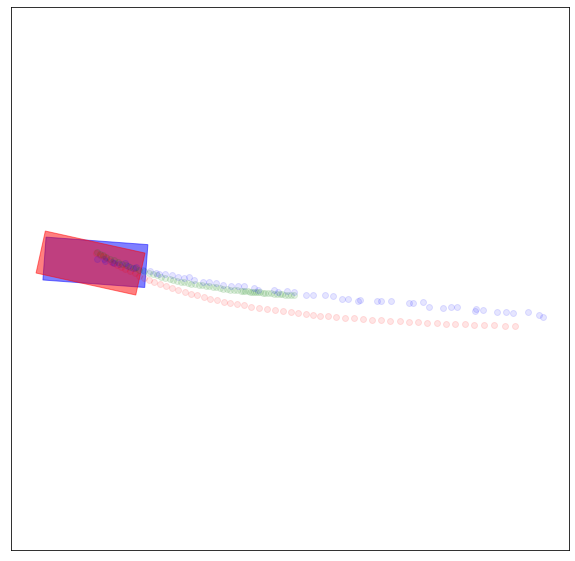}} 
{\includegraphics[width=.24\linewidth,clip]{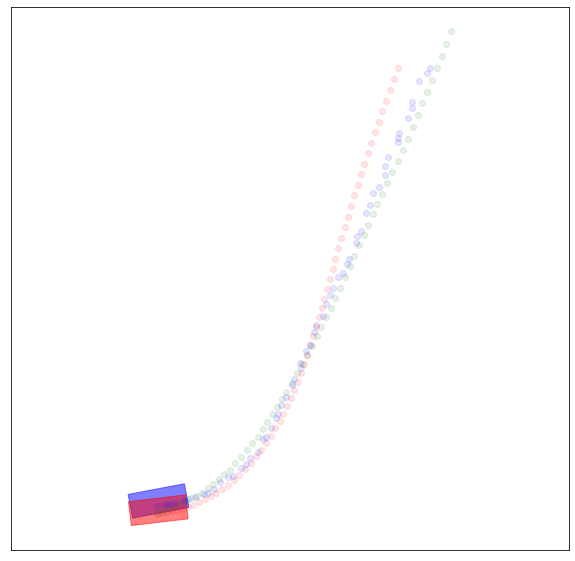}}
{\includegraphics[width=.24\linewidth,clip]{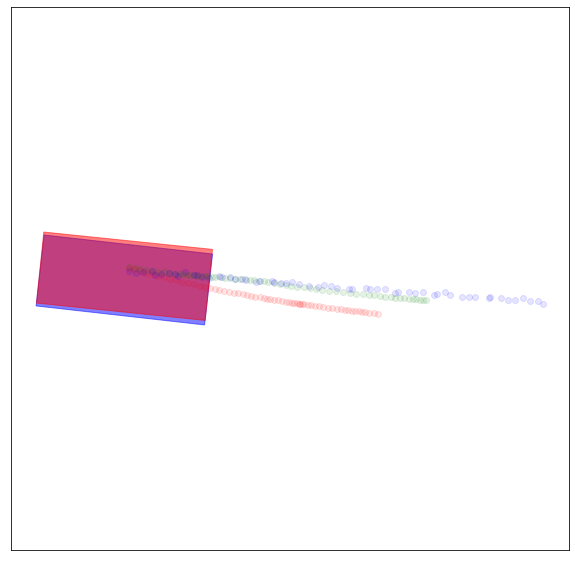}} 
\end{minipage}
\caption{Qualitative results comparing GoalNet-1T and PTNet-1T. We can see that PTNet's output is always smooth, while GoalNet's can be noisy and physically infeasible.}
\label{fig:qualitative}
\end{figure}

Next, we compare our method to GoalNet on the feasibility metrics described in Section \ref{sec:feasibility-metrics}. We omit a comparison with other baselines on these metrics because GoalNet is the strongest baseline in terms of trajectory error and it provides the most direct comparison for evaluating the impact of our structured ``robotics'' layer on the dynamic feasibility of the resulting trajectories. We report feasibility results on the NuScenes dataset. In this experiment, we trained GoalNet and PTNet using three different choices of temporal modes, denoted by 1T,2T,3T. For each model and for each metric, we report the fraction of total trajectories violating the metric. We count one trajectory as one violation regardless of the number of waypoints violating the requirement. The result is shown in Table \ref{tab:results-feasibility}. We observe that all variants of PTNet perform significantly better than all variants of GoalNet across all metrics. Specifically, because of the way PTNet is set up, curvature and traversal acceleration requirements are guaranteed to be satisfied. We confirm quantitatively that PTNet shows 0\% violations in these categories. We also observe that as we increase the number of temporal modes, i.e. increasing the number of trajectories produced per goal, GoalNet tends to produce more trajectories violating physical realism while the PTNet numbers stay relatively constant or decrease slightly. We also show qualitative plots in Figure \ref{fig:qualitative}. Qualitatively, we observe that PTNet's trajectory outputs, although sometimes further away from the ground truth in terms of L2 distance, are always smooth. In contrast, GoalNet is slightly more accurate in terms of average trajectory error metrics, but its trajectories can be noisy and non-smooth.

\begin{table}[tbp]
\begin{center}
\resizebox{\textwidth}{!}{
\begin{tabular}{lccccc}
\toprule
Method & Curvature & Lateral Speed & Centripetal Accel & Min Traversal Accel & Max Traversal Accel  \\
 & Violations & Violations (\%) & Violations (\%) &  Violations (\%) & Violations (\%) \\ 
\midrule
  GoalNet-1T & 65.16 & 0.43 & 16.64 & 0.06 & 0.76 \\
  GoalNet-2T & 68.80 & 4.41 & 22.34 & 1.90 & 3.23 \\
  GoalNet-3T & 69.02 & 6.71 & 25.86 & 3.60 & 5.64 \\
  PTNet-1T & 0  & 0.27 & 9.49 & 0 & 0 \\
  PTNet-2T & 0  & 0.27 & 9.21 & 0 & 0 \\
  PTNet-3T & 0  & 0.27 & 7.90 & 0 & 0 \\
  GroundTruth & 0 & 0 & 0 & 0 & 0\\
\bottomrule
\end{tabular}
}
\vspace{.2cm}
\caption{Trajectory feasibility metrics comparison on the public NuScenes dataset.}
\vspace{-.5cm}
\label{tab:results-feasibility}
\end{center}
\vspace{-.2cm}
\end{table}

\begin{figure}
\vspace{-2mm}
    \centering
    \includegraphics[width=0.7\linewidth,clip]{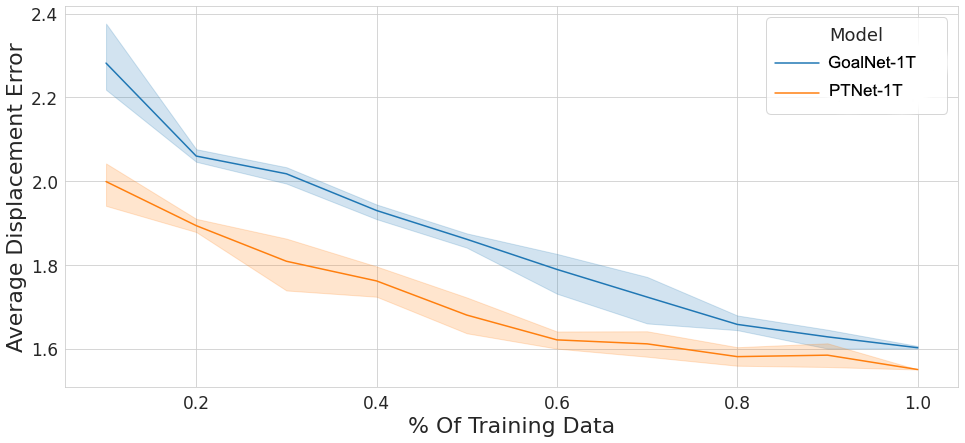}
    \caption{A comparison of the sample efficiency of PTNet-1T and GoalNet-1T. We plot the best-matching average displacement error against percentage of training data. We observe that PTNet can be up to 2 times more sample efficient to reach the same level of performance.}
    \label{fig:sample}
\vspace{-2mm}
\end{figure}

Lastly, we experiment on sample efficiency with GoalNet-1T and PTNet-1T. We set up the experiment by evaluating each model when trained on X\% of the training set and evaluating on the entire test set. For each value of X, we use four different random seeds to randomly select X\% of the training logs. We train both models on this exact set of sampled logs and evaluate on the test set. In this experiment, we report the performance of GoalNet-1T and PTNet-1T using the best matching average displacement error. Results are shown in Figure \ref{fig:sample}. We can see that because PTNet can leverage the distilled knowledge about the world through the dynamics equation update, the model requires 2X less data in order to converge to the same level of performance as GoalNet. On the other hand, GoalNet needs to learn physics from scratch, which requires a lot more data.

Overall, we observe that the addition of a robotics layer helps PTNet produce smooth and dynamically feasible trajectories while maintaining state of the art level performance on trajectory error metrics and requiring 2X less data.
\vspace{-.1cm}
\section{Conclusion}
\vspace{-1mm}
In this work, we introduce PTNet which combines a machine-learning-based trajectory prediction model with a structured path-tracking algorithm. Specifically, we introduce a differentiable path-tracking (PT) layer that can be added to existing architectures while still allowing end-to-end training. PTNet models and executes one or more acceleration profiles per path using a differentiable Pure Pursuit path tracker. The final output is a set of multi-modal spatial-temporal trajectories. By leveraging the power of learning from data as well as the structure embedded in classical physics-based robotics methods, PTNet is able to significantly improve on physical realism of the predicted trajectories while maintaining state-of-the-art performance on key trajectory error metrics and requiring two times less data. We view this as a small step towards a bigger idea, which is the fusion of ML and Robotics techniques for vehicle trajectory prediction.

\vspace{-3mm}
\newpage

{\small
\bibliography{egbib}
}

\newpage
\section{Appendix}
\vspace{-1mm}

\subsection{Error Metrics}
\subsubsection{Average Displacement Error}
We report the commonly used average displacement error (which we call avg DE, but is frequently abbreviated ADE) as a metric to quantify errors between predicted and ground-truth trajectories. This quantity is defined as the average $\ell_2$ distance between the predicted and ground truth future trajectories across all timesteps for all actors. 

\subsubsection{Average Along Track Error and Cross Track Error}
To further analyze prediction errors, we decompose errors into their along-track and cross-track components. These errors are calculated in the path-relative coordinate frame of the ground truth trajectory. We follow the methodology in \cite{zhang2020map} -- given the ground truth trajectory, $\tau_{xy}$, we re-sample $\tau_{xy}$ at a fixed spatial resolution of $\delta_\tau = 0.1$m, giving us the ground-truth path, $\rho^*_{xy}$. We then project the ground truth trajectory, $\tau_{xy}$, and predicted trajectory, $\hat{\tau}_{xy}$, onto the ground truth path $\rho^*_{xy}$. The projection decomposes the error into the along-track and cross track components. Given the along-track and cross-track representations of the ground truth trajectory, $\tau_{ac}$, and predicted trajectory, $\hat{\tau}_{ac}$, we calculate along-track error (ATE) and cross-track error (CTE) for a given timestep, $t$, as follows:

\begin{equation}
\textit{ATE} = \left|\hat{\tau}_{a}^t - \tau_{a}^t \right| \qquad \textit{CTE} = \left|\hat{\tau}_{c}^t \right|
\end{equation}

We report the average along-track error (avg ATE) and average cross-track error (avg CTE) by averaging over all timesteps across all actors.

\begin{figure}
    \centering
    \includegraphics[width=8.5cm]{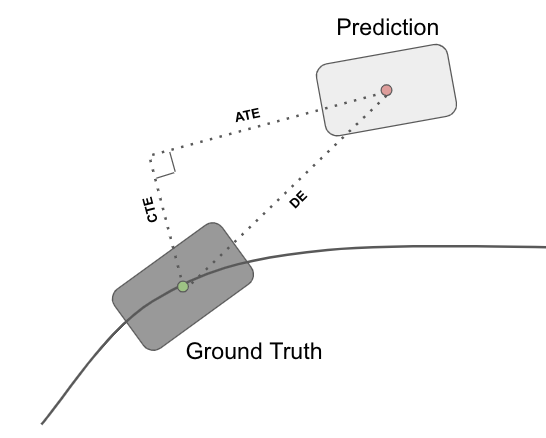}
    \caption{Illustration of the decomposition of displacement error (DE) into its along-track error (ATE) and cross-track error (CTE) components at a given timestep.}
    \label{fig:error}
\end{figure}

\end{document}